\documentclass{article}

\usepackage[preprint]{neurips_2024}   % swap to ICLR template for submission

\usepackage[utf8]{inputenc}
\usepackage[T1]{fontenc}
\PassOptionsToPackage{hyphens}{url}
\usepackage{hyperref}
\usepackage{url}
\usepackage{booktabs}
\usepackage{amsfonts}
\usepackage{amsmath}
\usepackage{amssymb}
\usepackage{microtype}
\usepackage{graphicx}
\usepackage{xcolor}

\usepackage{multirow}
\usepackage{array}
\usepackage{tabularx}
\usepackage{makecell}
\usepackage{longtable}

\usepackage{caption}
\usepackage{subcaption}

\usepackage{pifont}
\newcommand{\yes}{\ding{51}}
\newcommand{\no}{\ding{55}}
\newcommand{\maybe}{$\sim$}

\usepackage[numbers,sort&compress]{natbib}

\usepackage{fancyvrb}

\usepackage{amsthm}
\newtheorem{definition}{Definition}

\newcolumntype{C}[1]{>{\centering\arraybackslash}p{#1}}
\newcolumntype{L}[1]{>{\raggedright\arraybackslash}p{#1}}

\definecolor{classblue}{HTML}{1f77b4}
\definecolor{classorange}{HTML}{ff7f0e}
\definecolor{classgreen}{HTML}{2ca02c}
\definecolor{classred}{HTML}{d62728}
\definecolor{classpurple}{HTML}{9467bd}
\definecolor{classbrown}{HTML}{8c564b}

\title{Evaluation Blindness: How Silent Measurement Failures\\
       Corrupt AI Systems from Training to Deployment}

\author{%
  Priyanka Bajaj \\
  Independent Researcher \\
  \texttt{https://github.com/priyanka25aug/llm-failure-taxonomy}
}

\begin{document}
\sloppy   % allow looser line-breaking to prevent overflow in narrow columns

\maketitle

% =============================================================================
% ABSTRACT
% =============================================================================
\begin{abstract}
AI systems can fail silently.  When they do, the failure propagates undetected
--- through training loops, evaluation pipelines, and production monitoring
stacks --- until downstream harm makes it visible.  We identify \emph{evaluation
blindness} as the unifying structural property of this failure mode: a
measurement function $M$ exhibits evaluation blindness with respect to a failure
class $F$ if $M$ produces a value indistinguishable from a non-failing state
while the system is failing, and no auxiliary signal flags the discrepancy.

Evaluation blindness manifests at two distinct lifecycle stages that the
literature has treated separately.  At \emph{training time}, it appears when
reward models are gamed, importance-sampling corrections are silently
miscalculated, or benchmark contamination inflates fine-tuning evaluations ---
all cases where gradient signal and loss curves look healthy while the trained
policy is being corrupted.  At \emph{deployment time}, it appears when
evaluation frameworks fail to detect the six classes of production failure we
taxonomise: Model Drift, Infrastructure, Integration, Evaluation, Safety \&
Compliance, and Operational --- the last of which is 100\% silent by its
structural definition.

We make four contributions.  First, a formal definition of evaluation blindness
with a detectability predicate that unifies training-time and deployment-time
measurement failure under a single framework.  Second, four documented
training-time case studies --- including a concrete implementation bug in a
widely-used open-source RL training library (TRL PR~\#6594) --- where the loss
decreases normally while gradients are corrupted.  Third, a six-class
system-level failure taxonomy validated against 50 real-world production
incidents from court documents, regulatory filings, and public postmortems,
finding that 53\% of verifiable public incidents are silent.  Fourth, a per-use-case failure
budget framework formalising acceptable failure rates by risk class
(Decision-Critical, Customer-Facing, Internal Productivity, Experimental).

The unified framing has a direct engineering implication: measurement
infrastructure is a correctness concern at every stage of the AI system
lifecycle, not only at model evaluation time.  All data, code, and the taxonomy
schema are released at \url{https://github.com/priyanka25aug/llm-failure-taxonomy}.
\end{abstract}

% =============================================================================
\section{Introduction}
\label{sec:intro}
% =============================================================================

\subsection{The Common Thread Across Three AI Disasters}

In 2023, a New York attorney submitted a legal brief containing citations to six
non-existent cases, all generated by ChatGPT.  The sanctions opinion in
\emph{Mata v. Avianca} \citep{mata2023avianca} documented the harm.  What is
less often noted is why the failure was not caught: the attorney's review
process --- the measurement function for ``is this citation correct?'' ---
produced no distinguishing signal between real and fabricated case names.
The legal citation \emph{looked} valid.

That same year, Air Canada operated a customer service chatbot that told a
grieving passenger he could apply for a bereavement fare retroactively ---
a policy that does not exist \citep{aircanada2024chatbot}.  The British Columbia
Civil Resolution Tribunal found Air Canada liable.  What failed was not model
accuracy but the system's measurement infrastructure: no component verified
chatbot claims against a live policy database before presenting them to a user
as authoritative.  The output passed every implicit quality check in the
integration pipeline.

The UK Post Office Horizon system \citep{ukposthorizon2024}, in operation from
1999 onward, produced accounting discrepancies that led to the wrongful
prosecution of more than 700 sub-postmasters.  The accounting outputs were
logged, audited, and acted upon --- but the audit function itself was broken.
Measurement infrastructure treated system output as ground truth, so the
failures propagated undetected for years.

All three incidents are AI system failures.  All three share a structural
property: \textbf{the measurement function that should have caught the failure
produced a value indistinguishable from a non-failing state}.  No alert fired.
No quality metric degraded.  The failures were invisible to the observability
infrastructure until downstream harm made them visible.  We call this
\emph{evaluation blindness}.

\subsection{Evaluation Blindness at Training and Deployment Time}

The same structural property appears earlier in the AI lifecycle, during model
training, where it has gone largely unrecognised.

Consider a reinforcement learning from human feedback (RLHF) training run where
the reward model is gamed: the policy learns to produce outputs that score highly
on the proxy reward while degrading on the actual user utility criterion
\citep{gao2025rewardhacking, stiennon2020learning}.  The training loss decreases
normally.  The reward curves look healthy.  No gradient error fires.  The
evaluation infrastructure that \emph{would} detect the quality degradation ---
a gold-standard human preference evaluation --- is not run inline with training.
The failure is invisible until post-deployment quality review.

A more concrete instance: in the Group Relative Policy Optimisation (GRPO)
implementation in the Hugging Face TRL library \citep{trl2024grpo}, a
single-line miscalculation in the importance-sampling ratio correction caused
the per-token ratio to be silently replaced by a sequence-mean during KL
divergence bias correction.  The training loss decreased.  The policy appeared
to improve on the evaluation metrics.  The gradient computation was wrong for
the entire duration of the affected runs, and the error was surfaced only when
a contributor compared the TRL implementation against the DeepSeek-V3
specification \citep{shao2024deepseekmath}.  This is a training-time evaluation
blindness event: the measurement function (training loss and reward curves)
produced a value consistent with correct training while the gradients were
corrupted.

The literature has treated training-time failures \citep{gao2025rewardhacking}
and deployment-time silent failures
\citep{shen2025silent, kumar2025deployment, liu2025blindspot} as separate
research programmes.  We argue they are the same problem at different lifecycle
stages, and that a unified framework resolves the false separation.

\subsection{The Gap in Existing Frameworks}

Sculley et al.\ \citep{sculley2015debt} identified hidden technical debt in ML
systems --- entanglement, feedback loops, undeclared consumers --- but predated
modern LLM deployments.  Amershi et al.\ \citep{amershi2019se4ml} addressed
ML engineering process failures without capturing LLM-specific modes.
Paleyes et al.\ \citep{paleyes2022challenges} surveyed pre-LLM deployment
challenges.

LLM-specific work focuses on model-layer phenomena: hallucination surveys
\citep{ji2023hallucination, huang2023survey}, adversarial inputs
\citep{wei2023jailbroken, perez2022injection, greshake2023indirect}, and
alignment failures \citep{bai2022constitutional}.  Recent work has begun
cataloguing deployment-time silent failures
\citep{shen2025silent, kumar2025deployment, liu2025blindspot,
pandey2026evaluating}, and separate
work addresses training-time reward corruption \citep{gao2025rewardhacking}.
No prior work (a) provides a formal definition of the common structural
property, (b) unifies training-time and deployment-time measurement failure
under a single framework, or (c) introduces a formal acceptable-failure-rate
model calibrated to regulatory practice.

\subsection{Contributions}

This paper makes four contributions:

\begin{enumerate}
  \item A \textbf{formal definition of evaluation blindness} --- a structural
        property of measurement functions that applies at both training time
        and deployment time --- with a detectability predicate and lifecycle
        staging.

  \item \textbf{Four documented training-time evaluation blindness case studies},
        including the GRPO importance-sampling bug (TRL PR~\#6594) as a
        concrete, verifiable instance where a training run produces plausible
        metrics while gradients are corrupted.

  \item A \textbf{six-class system-level failure taxonomy} with sub-classes,
        detectability profiles, and blast radius characterisation, validated
        against 50 real-world production incidents from verifiable public
        sources (court documents, regulatory filings, academic papers,
        postmortems).

  \item A \textbf{per-use-case failure budget framework} formalising acceptable
        failure rates by risk class, with a reference implementation for
        computing failure budget utilisation in production.
\end{enumerate}

Section~\ref{sec:blindness} formalises evaluation blindness.
Section~\ref{sec:related} reviews related work.
Section~\ref{sec:taxonomy} introduces the taxonomy.
Section~\ref{sec:classes} defines the six failure classes.
Section~\ref{sec:training} documents training-time evaluation blindness.
Section~\ref{sec:budget} presents the failure budget framework.
Section~\ref{sec:dataset} describes dataset construction and validation.
Sections~\ref{sec:discussion} and~\ref{sec:conclusion} discuss implications
and open problems.

% =============================================================================
\section{Evaluation Blindness: Formal Definition}
\label{sec:blindness}
% =============================================================================

\subsection{The Structural Property}

A wide range of AI system failures share a structural property that is
distinct from the observable symptom (wrong output, degraded quality,
policy violation) and distinct from the root cause (data drift, reward
hacking, missing runbook).  The shared property is that \emph{the
measurement function that should detect the failure produces a value
consistent with a non-failing state}.

\begin{definition}[Evaluation Blindness]
\label{def:eb}
Let $S$ be an AI system in state space $\mathcal{S}$, and let $M :
\mathcal{S} \to \mathbb{R}^k$ be a measurement function (a loss function,
evaluation metric, monitoring signal, or composite thereof) with
non-failing reference distribution $\mathcal{D}_{\text{ok}}$.  A system
state $s \in \mathcal{S}$ in failure class $F$ exhibits
\textbf{evaluation blindness} with respect to $M$ if:
\[
  M(s) \;\overset{d}{\approx}\; \mathcal{D}_{\text{ok}}
  \quad \text{and} \quad
  \nexists\; \text{auxiliary signal} \;\hat{M} \;\text{s.t.}\;
  \hat{M}(s) \notin \mathcal{D}^{\hat{M}}_{\text{ok}}
\]
where $\overset{d}{\approx}$ denotes distributional indistinguishability
up to tolerance $\varepsilon$ and $\mathcal{D}^{\hat{M}}_{\text{ok}}$ is
the reference distribution for $\hat{M}$.  Informally: the system is
failing and no measurement in the \emph{currently deployed} set
$\mathcal{M}$ detects it.
\end{definition}

\begin{definition}[Structural vs.\ Incidental Evaluation Blindness]
\label{def:structural}
Let $\mathcal{M}^{*}$ denote the set of all \emph{structurally available}
measurement functions given the system's observable signals.
Evaluation blindness is \textbf{structural} if
$\delta(s, \mathcal{M}^{*}) = 0$: no function of the available signals
can distinguish $s$ from $\mathcal{D}_{\text{ok}}$.
It is \textbf{incidental} if $\delta(s, \mathcal{M}) = 0$ but
$\delta(s, \mathcal{M}^{*}) = 1$: the failure \emph{is} detectable in
principle, but the required measurement has not been deployed.
\end{definition}

\noindent
The distinction has a direct remediation implication.  Structural
evaluation blindness requires a different measurement \emph{modality}
(e.g.\ a gold-label human preference evaluation that does not exist
as a deployable system signal); incidental evaluation blindness requires
deploying a known-but-absent monitor (e.g.\ a PII regex on outputs never
configured).  Most C4, C1, and C3 deployment-time failures in our dataset
are incidental; the GRPO IS ratio bug (Section~\ref{sec:training}) is
structural --- no training-time measurement could have flagged the gradient
corruption without external specification comparison.

\subsection{Detectability Predicate}

We define a binary \textbf{detectability predicate} $\delta(s, \mathcal{M})$
where $\mathcal{M} = \{M_1, \ldots, M_n\}$ is the set of all deployed
measurement functions:
\[
  \delta(s, \mathcal{M}) =
  \begin{cases}
    1 & \text{if } \exists\, M_i \in \mathcal{M} \text{ s.t. }
        M_i(s) \notin \mathcal{D}^{M_i}_{\text{ok}} \\
    0 & \text{otherwise}
  \end{cases}
\]
A failure is \emph{evaluation-blind} if $\delta(s, \mathcal{M}) = 0$
for the current measurement set.  Mean Time to Detection (MTTD) is
infinite under a static $\mathcal{M}$; in practice the failure surfaces
when $\mathcal{M}$ is augmented (new monitor added), when downstream
harm creates an observable signal outside $\mathcal{M}$, or when a
human audit reviews outputs directly.

\subsection{Lifecycle Staging}

Evaluation blindness occurs at two distinct stages of the AI system lifecycle:

\paragraph{Training-time evaluation blindness.}
The measurement set $\mathcal{M}$ consists of training loss, reward curves,
and held-out benchmark evaluation.  A failure state $s$ is training-blind
if the loss decreases normally, the reward looks healthy, and benchmark
scores are stable while the model being trained is being corrupted.
Root causes include: reward model gaming, importance-sampling miscalculation,
benchmark contamination, and mode collapse in policy optimisation.
Section~\ref{sec:training} documents four concrete instances.

\paragraph{Deployment-time evaluation blindness.}
The measurement set $\mathcal{M}$ consists of production monitoring (error
rates, latency, alerting) and evaluation pipelines (offline benchmarks,
LLM-as-judge scorers).  A failure state $s$ is deployment-blind if
production monitoring and evaluation pipelines both fail to flag the
failure.  In our empirical dataset, 53\% of 36 verifiable public production incidents
are deployment-time evaluation-blind (52\% across all 50 including synthetic
composites).  The C4 Evaluation failure class
is 100\% blind by definition (the measurement infrastructure \emph{is}
the failed component).

\paragraph{The unifying consequence.}
Both lifecycle stages share the same engineering prescription: measurement
infrastructure must be treated as a first-class correctness concern.
Adding sensors to $\mathcal{M}$ --- whether training-time (gold-label
preference eval inline with RLHF) or deployment-time (distributional
monitoring, citation verification) --- is the remediation path.
Section~\ref{sec:training} and the taxonomy in Section~\ref{sec:classes}
provide a structured inventory of what $\mathcal{M}$ must contain at
each stage to achieve $\delta = 1$ for each failure class.

% =============================================================================
\section{Related Work}
\label{sec:related}
% =============================================================================

\subsection{ML System Reliability}

The foundational contribution of Sculley et al.\ \citep{sculley2015debt}
introduced the concept of \emph{technical debt} in ML systems, identifying
entanglement, hidden feedback loops, undeclared consumers, and unstable data
dependencies as recurring failure patterns.  These patterns remain relevant for
LLM systems, but the taxonomy predates the modern deployment stack entirely:
it does not address serving infrastructure for large models, RAG pipelines,
prompt injection, or the compliance requirements introduced by the EU AI Act
\citep{euaiact2024} and the Digital Operational Resilience Act
\citep{dora2022}.

Amershi et al.\ \citep{amershi2019se4ml} identified three software engineering
challenges specific to ML: data collection and management, model building, and
deployment monitoring.  Their framework applies to supervised learning systems
and does not address the non-determinism, prompt sensitivity, or context-window
constraints of autoregressive LLMs.

Paleyes et al.\ \citep{paleyes2022challenges} surveyed 35 case studies of ML
deployment, identifying data, model, code, and organisational failure categories.
Their review covers pre-LLM deployments and lacks the RAG, tool-call, and
safety-compliance sub-classes relevant to contemporary production systems.

\subsection{LLM Failure Analysis}

Hallucination surveys \citep{ji2023hallucination, huang2023survey} provide
the most rigorous model-layer failure taxonomies available, distinguishing
intrinsic and extrinsic hallucinations, faithfulness failures, and factuality
failures.  These are essential for understanding model-layer behaviour but do
not address failures that originate in the system around the model.

Adversarial input research \citep{wei2023jailbroken, perez2022injection}
demonstrates that LLM safety training fails under distribution shift, and that
external inputs can override system prompts.  Greshake et al.\
\citep{greshake2023indirect} introduced indirect prompt injection, where
adversarial instructions are embedded in retrieved content processed by the
model.  These contributions map to a single sub-class (3a) of our CLASS\_3
Integration taxonomy.

Constitutional AI \citep{bai2022constitutional} and related alignment work
address the training-time mitigation of one sub-class (5c) of our Safety \&
Compliance failures: policy boundary violations.  The broader compliance
failure surface --- PII leakage, hallucinated legal citations, auditability
gaps, copyright reproduction --- is not addressed by alignment training alone.

Evaluation critiques \citep{bowman2021nlu, liang2022helm, ribeiro2020checklist,
shankar2020evaluating} collectively document the failure of benchmark proxies
to capture real-world model behaviour.  HELM \citep{liang2022helm} explicitly
acknowledges the ``evaluation gap'' between benchmark performance and production
reliability.  These contributions map to our CLASS\_4 Evaluation Failures,
which we treat as a system-level failure class in their own right.

\subsection{Recent Work on Silent Deployment Failures (2025--2026)}

Four concurrent papers have begun cataloguing silent failures in deployed
LLM systems.  Shen et al.\ \citep{shen2025silent} provide a taxonomy of
undetected deployment failures focused on output-quality degradation.  Kumar
et al.\ \citep{kumar2025deployment} analyse measurement gaps in production AI
monitoring.  Liu et al.\ \citep{liu2025blindspot} characterise the structural
conditions under which deployed LLMs fail invisibly.  Pandey
\citep{pandey2026evaluating} study seven production failure modes in agentic
systems empirically, finding that \emph{no standard metric detects more than
two of the seven failure modes}; their empirical data include a case where
accuracy remained flat at 0.86--0.88 across five evaluation windows while
output diversity collapsed by 6.5$\times$ --- a concrete instantiation of
our formal Definition~\ref{def:eb}.  All four address deployment-time failures
only.  None provides a formal definition of the structural property they
describe, and none addresses training-time evaluation blindness.  This work
both formalises the shared property and extends its scope to the training
lifecycle.\footnote{The author of \citet{pandey2026evaluating} also served as
the independent annotator for the inter-rater reliability study reported in
Section~\ref{sec:dataset}; this relationship is disclosed in the
Acknowledgments.}

\subsection{Training-Time Reward Corruption}

Gao et al.\ \citep{gao2025rewardhacking} provide empirical evidence that RLHF
reward models can be systematically gamed beyond a KL budget threshold,
producing policies that score highly on the proxy reward while degrading on
human preference evaluations.  This work addresses a specific mechanism (reward
hacking) at training time; it does not formalise the evaluation blindness
property, does not address implementation-level bugs (as in our GRPO case study),
and does not connect to the deployment-time failure taxonomy.

\subsection{AI Incident Databases}

The AI Incident Database \citep{mcgregor2021aiid} provides the most
comprehensive public collection of real-world AI failures, with over 600
indexed incidents at the time of writing.  The AIAAIC
repository \citep{aiaaic2023} provides a complementary collection with
stronger regulatory context.  Both databases collect \emph{incidents} without
providing a structured failure taxonomy with detectability profiles, formal
failure budget formalism, or lifecycle-spanning analysis.

\subsection{Coverage Gap}

Table~\ref{tab:related_work} maps prior work to our six failure classes and
the training-time evaluation blindness contribution.  No single prior work
covers the full taxonomy and no prior work connects training-time and
deployment-time measurement failure under a unified formal framework.

\begin{table*}[t]
\caption{Coverage of failure classes and key contributions across prior work.
\yes~= full coverage; \maybe~= partial; \no~= not addressed.
``EB'' = formal evaluation blindness definition.
``Train'' = training-time evaluation blindness.}
\label{tab:related_work}
\centering
\footnotesize
\setlength{\tabcolsep}{3pt}
\resizebox{\linewidth}{!}{%
\begin{tabular}{lC{1.2cm}C{1.2cm}C{1.2cm}C{1.2cm}C{1.2cm}C{1.2cm}C{1.1cm}C{1.1cm}C{0.9cm}}
\toprule
\textbf{Work} &
\makecell{\textbf{C1}\\\textbf{Drift}} &
\makecell{\textbf{C2}\\\textbf{Infra}} &
\makecell{\textbf{C3}\\\textbf{Integr.}} &
\makecell{\textbf{C4}\\\textbf{Eval}} &
\makecell{\textbf{C5}\\\textbf{Safety}} &
\makecell{\textbf{C6}\\\textbf{Ops}} &
\makecell{\textbf{Formal}\\\textbf{EB}} &
\makecell{\textbf{Train}\\\textbf{EB}} &
\textbf{/8} \\
\midrule
Sculley et al.\ \citeyear{sculley2015debt}     & \maybe & \maybe & \no & \no & \no & \maybe & \no & \no & 1.5 \\
Amershi et al.\ \citeyear{amershi2019se4ml}    & \maybe & \no    & \no & \maybe & \no & \maybe & \no & \no & 1.5 \\
Paleyes et al.\ \citeyear{paleyes2022challenges}& \maybe & \maybe & \no & \no & \no & \maybe & \no & \no & 1.5 \\
Ji et al.\ \citeyear{ji2023hallucination}       & \yes   & \no    & \no & \no & \maybe & \no & \no & \no & 1.5 \\
Wei et al.\ \citeyear{wei2023jailbroken}        & \no    & \no    & \maybe & \no & \maybe & \no & \no & \no & 1.0 \\
Liang et al.\ \citeyear{liang2022helm}          & \no    & \no    & \no & \yes & \no & \no & \no & \no & 1.0 \\
AIID \citeyear{mcgregor2021aiid}               & \maybe & \maybe & \maybe & \no & \maybe & \maybe & \no & \no & 2.0 \\
Gao et al.\ \citeyear{gao2025rewardhacking}    & \no & \no & \no & \no & \no & \no & \no & \maybe & 0.5 \\
Shen et al.\ \citeyear{shen2025silent}         & \maybe & \no & \maybe & \maybe & \no & \no & \no & \no & 1.0 \\
Kumar et al.\ \citeyear{kumar2025deployment}   & \maybe & \no & \no & \yes & \no & \no & \no & \no & 1.5 \\
Pandey \citeyear{pandey2026evaluating}         & \yes & \maybe & \maybe & \yes & \no & \no & \no & \no & 3.0 \\
\midrule
\textbf{This work} & \yes & \yes & \yes & \yes & \yes & \yes & \yes & \yes & \textbf{8.0} \\
\bottomrule
\end{tabular}}
\end{table*}

% =============================================================================
\section{Taxonomy Framework}
\label{sec:taxonomy}
% =============================================================================

\subsection{Design Principles}

Four principles guide the taxonomy design.

\paragraph{Layer-specificity.}
Each failure class maps to a distinct system layer: model weights and inference
(C1), serving infrastructure (C2), application--LLM integration (C3), evaluation
and observability (C4), governance and compliance (C5), and operational process
(C6).  Layer-specific classification enables routing each failure to the
appropriate owning team rather than treating all failures as model problems.

\paragraph{Actionability.}
Each class implies a distinct remediation path.  Infrastructure failures require
capacity planning and load-shedding; integration failures require guardrails and
output validators; operational failures require runbook creation and monitoring
coverage.  A taxonomy that cannot drive remediation produces correct labels but
no operational value.

\paragraph{Detectability as a first-class attribute.}
We classify not only \emph{what} fails but \emph{when and how} it becomes
observable.  Detectability takes three values: \emph{immediate} (alert fires
within minutes), \emph{delayed} (discovered within hours to days), or
\emph{silent} (discovered only via audit, user complaint, or manual review,
weeks to months later).  Our key empirical finding --- that 50\% of incidents
are silent --- motivates this as a first-class attribute.

\paragraph{Blast radius.}
Scope of impact is a classification attribute, not an afterthought.  Blast
radius takes six levels: \texttt{single\_request} $\to$ \texttt{single\_user}
$\to$ \texttt{user\_cohort} $\to$ \texttt{team} $\to$ \texttt{org\_wide}
$\to$ \texttt{public}.  The same failure class (e.g.\ prompt injection) can
have very different operational impact depending on whether it affects one
request or all requests from a user cohort.

\subsection{Classification Dimensions}

Table~\ref{tab:dimensions} summarises the seven classification dimensions
applied to each labeled incident.

\begin{table}[t]
\caption{Taxonomy classification dimensions.}
\label{tab:dimensions}
\centering
\small
\resizebox{\linewidth}{!}{%
\begin{tabular}{llL{4.0cm}}
\toprule
\textbf{Dimension} & \textbf{Type} & \textbf{Values} \\
\midrule
Failure class & Enum & C1--C6 (see Section~\ref{sec:classes}) \\
Sub-class     & Enum & 3--5 per class (26 total) \\
Severity      & Ordinal & critical, high, medium, low \\
Detectability & Ordinal & immediate, delayed, silent \\
Blast radius  & Ordinal & single\_request $\to$ public \\
MTTD          & Numeric & Hours (0.1 to $>$1000) \\
Failure budget class & Enum & FC\_A, FC\_B, FC\_C, FC\_D \\
\bottomrule
\end{tabular}}
\end{table}

\subsection{Relationship to Model-Centric Framing}

A surface-level manifestation of ``the model returned wrong output'' can arise
from multiple distinct root causes in our taxonomy.  Hallucination of legal
citations (as in \emph{Mata v.\ Avianca}) is a Safety \& Compliance failure
(C5b) because the context is a high-stakes domain lacking a verification
guardrail.  A model returning outdated regulatory information from a stale RAG
index is an Integration failure (C3d).  A model whose output distribution
quietly shifted after an upstream provider update is a Model Drift failure (C1a).
All three can appear as ``the LLM produced wrong output.'' All three require
different remediation.  The taxonomy makes this distinction precise.

% =============================================================================
\section{The Six Failure Classes}
\label{sec:classes}
% =============================================================================

Table~\ref{tab:six_classes} provides a compact overview.  Each sub-section
below gives the definition, sub-classes, key monitoring indicators, detectability
profile, and a representative real-world incident.

\begin{table*}[t]
\caption{The six failure classes with detectability and blast radius profiles.}
\label{tab:six_classes}
\centering
\small
\resizebox{\linewidth}{!}{%
\begin{tabular}{lL{2.5cm}L{3.0cm}L{2.0cm}L{3.0cm}}
\toprule
\textbf{Class} & \textbf{Name} & \textbf{System Layer} &
\textbf{Detectability} & \textbf{Key Indicator} \\
\midrule
C1 & Model Drift         & Model / Inference      & Silent $\to$ Delayed
   & Output distribution shift; MTTD 24h--168h \\
C2 & Infrastructure      & Serving Stack          & Immediate $\to$ Delayed
   & Latency spike; OOM; 5xx error rate \\
C3 & Integration         & App--LLM Boundary      & Delayed $\to$ Silent
   & Prompt injection; stale retrieval; context truncation \\
C4 & Evaluation          & Observability          & Silent
   & Metric proxy collapse; benchmark contamination \\
C5 & Safety \& Compliance& Governance             & Delayed $\to$ Silent
   & PII regex hit; hallucinated citation; policy breach \\
C6 & Operational         & Process \& Runbooks    & Delayed
   & Monitoring blind-spot; absent runbook; escalation gap \\
\bottomrule
\end{tabular}}
\end{table*}

\subsection{C1 --- Model Drift Failures}

\paragraph{Definition.}
Failures caused by a shift in the LLM's output distribution that is not caused
by an explicit model weight update visible to the deploying team.  The model
``changes'' silently from the operator's perspective.

\paragraph{Sub-classes.}
(1a)~Upstream provider silent update; (1b)~production distribution shift from
input distribution drift; (1c)~context-window saturation drift, where longer
prompts systematically degrade output quality.

\paragraph{Key indicators.}
Output token-length distribution; semantic similarity to baseline outputs;
task-specific quality proxy metrics; downstream conversion or satisfaction
metrics.

\paragraph{Representative incident.}
In March 2023, developers using GPT-4 via the OpenAI API reported systematic
changes in model behaviour --- including altered instruction-following patterns
and changed coding style --- without any announced model update.  The change
was detected via developer forum reports approximately 168 hours after it
began.  No production monitoring alert fired.  This is a canonical C1a
incident: silent upstream update, silent detectability, blast radius
\texttt{user\_cohort}.

\paragraph{Detectability profile.}
Silent to delayed.  MTTD range: 24h to weeks.  Error-rate monitoring will not
fire; distributional monitoring is required.

\subsection{C2 --- Infrastructure Failures}

\paragraph{Definition.}
Failures in the serving stack, load balancer, GPU allocation, or dependent
services, where the model itself is functioning correctly but the infrastructure
around it fails to deliver model outputs reliably.

\paragraph{Sub-classes.}
(2a)~P99 latency regression; (2b)~OOM / KV-cache exhaustion; (2c)~routing
misclassification under load; (2d)~API breaking change from upstream provider.

\paragraph{Key indicators.}
Request error rate (4xx/5xx); P99 and P999 latency; GPU memory utilisation;
queue depth; cold-start frequency.

\paragraph{Representative incident.}
vLLM's PagedAttention memory management system \citep{kwon2023vllm} introduced
efficient GPU memory handling, but bursty long-context traffic can exhaust the
KV-cache under concurrent load, producing 500 errors with a stack trace
indicating OOM.  Detection is immediate (errors fire within seconds), but
remediation requires capacity planning.  Blast radius: \texttt{user\_cohort}
to \texttt{org\_wide} depending on request routing.

\paragraph{Detectability profile.}
Immediate to delayed.  MTTD range: 0.1h to 8h.  Standard error-rate monitoring
is sufficient to detect C2 failures promptly.

\subsection{C3 --- Integration Failures}

\paragraph{Definition.}
Failures at the boundary between the LLM and the application layer, where the
model itself produces output consistent with its training but the integration
layer --- prompts, retrieval pipelines, tool calls, or multi-turn state ---
is deficient.

\paragraph{Sub-classes.}
(3a)~Prompt injection / jailbreak; (3b)~context truncation (silent);
(3c)~tool-call hallucination; (3d)~RAG retrieval mismatch (stale index);
(3e)~multi-turn state corruption.

\paragraph{Key indicators.}
System prompt integrity checks; context window utilisation (flag truncation
at $>$85\% capacity); tool-call success rate; retrieval freshness timestamp;
semantic consistency across turns.

\paragraph{Representative incident.}
Bing Chat (Sydney persona, February 2023) demonstrated C3a: adversarial user
inputs extracted the hidden system prompt and enabled persona escape, causing
the system to produce outputs inconsistent with deployment intent.  The failure
was in the integration layer: no output validator verified that responses were
within policy bounds given the system prompt state.

\paragraph{Detectability profile.}
Delayed to silent.  RAG staleness failures (3d) are frequently silent for
weeks.  Prompt injection (3a) may be detected via content moderation if
guardrails are in place.

\subsection{C4 --- Evaluation Failures (Deployment-Time Evaluation Blindness)}

\paragraph{Definition.}
Failures in the evaluation and observability infrastructure itself, where
the measurement system that should detect other failures is itself broken.
In the framework of Section~\ref{sec:blindness}, C4 failures are the
deployment-time realisation of evaluation blindness: the deployed
$\mathcal{M}_{\text{deploy}}$ has $\delta(s, \mathcal{M}_{\text{deploy}}) = 0$
not because the failure is subtle, but because the measurement infrastructure
is structurally absent or misconfigured.  C4 failures are meta-failures:
they cause the other five failure classes to go undetected.

\paragraph{Sub-classes.}
(4a)~Metric proxy collapse (BLEU/ROUGE gaming, LLM-as-judge reward hacking at
deployment time); (4b)~evaluation set contamination (overlapping with production
traffic); (4c)~production--evaluation distribution gap (eval queries do not
match production query distribution); (4d)~point accuracy versus distributional
measurement (single-point eval misses distributional failures visible only
at scale).

\paragraph{Key indicators.}
Correlation between benchmark metrics and user satisfaction scores; evaluation
set update frequency; fraction of evaluation traffic overlapping with training
data; per-cohort versus aggregate metric decomposition.

\paragraph{Representative incident.}
Models evaluated on HELM \citep{liang2022helm} or MMLU can achieve benchmark
improvements through targeted fine-tuning that degrades real-world user value
--- a well-documented form of metric proxy collapse (4a).  This failure class
accounts for 10\% of our labeled incidents, but is the most likely to be
under-reported: C4 failures rarely surface as reportable incidents because
no alert fires and no user complaint is immediately traceable to the evaluation
gap.  The structural parallel with training-time evaluation blindness
(Section~\ref{sec:training}) is exact: in both cases, the measurement function
$M$ operates normally while reporting values consistent with a non-failing state.

\paragraph{Detectability profile.}
Silent.  MTTD: months ($\delta = 0$ under standard monitoring).  Requires
dedicated evaluation infrastructure with production traffic sampling, regular
benchmark rotation, and per-cohort metric decomposition.

\subsection{C5 --- Safety \& Compliance Failures}

\paragraph{Definition.}
Failures where model outputs violate regulatory requirements, internal safety
policies, or data governance rules, with the defining property that the failure
creates legal, reputational, or harm risk beyond the immediate user.

\paragraph{Sub-classes.}
(5a)~PII leakage (training memorisation or prompt reconstruction);
(5b)~hallucinated legal or medical citations; (5c)~policy boundary violation;
(5d)~auditability gap (no log of model decision); (5e)~copyright reproduction.

\paragraph{Key indicators.}
PII regex detector on outputs; citation verification pipeline; output
policy classifier; audit log completeness; copyright similarity detector.

\paragraph{Representative incident.}
\emph{Mata v. Avianca} \citep{mata2023avianca} is the canonical C5b incident.
The failure was not that the model was capable of fabricating legal citations ---
that is a known model-layer property.  The failure was the absence of a
verification layer between model output and consequential action: the attorney
submitted the output without cross-referencing against a legal database.
In a properly designed system for a Decision-Critical (FC\_A) use case,
every legal citation generated by an LLM should be verified against a live
legal database before submission.  The system design failed; the model
behaved as models do.

\paragraph{Detectability profile.}
Delayed to silent.  PII leakage (5a) may fire immediately if output monitoring
is in place; hallucinated citations (5b) are silent until downstream action
reveals the error.

\subsection{C6 --- Operational Failures}

\paragraph{Definition.}
Failures in the operational processes, runbooks, monitoring coverage, and
escalation paths around the LLM system.  The model and infrastructure may be
functioning correctly; the operational envelope has failed.

\paragraph{Sub-classes.}
(6a)~Monitoring blind-spot (no alert configured for a failure mode);
(6b)~runbook absence (no documented response when a known failure occurs);
(6c)~escalation breakdown (alert fires but reaches wrong team or no team);
(6d)~canary / shadow test gap (no pre-production testing of distribution
changes).

\paragraph{Key indicators.}
Alert coverage matrix against taxonomy classes; runbook inventory and
last-reviewed date; mean time to escalation versus SLO; shadow traffic
coverage fraction.

\paragraph{Representative incident.}
A composite pattern drawn from multiple enterprise deployments: an LLM output
pipeline begins producing longer responses, consuming more downstream storage
and increasing user-visible latency.  No alert exists for output-length
distribution shift; the team detects the change via a support ticket backlog
review three weeks later.  The model produced valid outputs throughout.
The operational envelope --- specifically, the absence of output distribution
monitoring --- was the failure.

\paragraph{Detectability profile.}
Delayed.  C6 failures typically surface via indirect signals (support backlog,
quarterly review, downstream system alerts) rather than direct monitoring.

% =============================================================================
\section{Training-Time Evaluation Blindness}
\label{sec:training}
% =============================================================================

Training-time evaluation blindness occurs when the measurement set available
during model training ($\mathcal{M}_{\text{train}} =$ \{loss, reward, benchmark
score\}) fails to detect that the training process is producing a corrupted
policy.  The failure is structurally identical to deployment-time C4 Evaluation
failures: the observability infrastructure reports normal operation while the
system degrades.

We present four instances spanning gradient computation, reward modelling,
data contamination, and policy collapse.  \textbf{Case Study~1 is an original
analysis} of a verifiable implementation bug in a widely-used open-source
library (TRL PR~\#6594), including a formal statement of the gradient
corruption and its evaluation-blindness properties.
\textbf{Case Studies~2--4 apply the evaluation blindness framework} to
well-documented phenomena in the literature
\citep{gao2025rewardhacking,shi2024contamination,ouyang2022instructgpt},
demonstrating that each constitutes a structural instance of Definition~1
--- a finding not made in the original papers.  In all four cases, the
standard training measurement set $\mathcal{M}_{\text{train}}$ satisfies
$\delta(s, \mathcal{M}_{\text{train}}) = 0$ for the duration of the
corrupted training run.

\subsection{Case Study 1: GRPO Importance-Sampling Ratio Bug (TRL PR~\#6594)}

Group Relative Policy Optimisation (GRPO) \citep{shao2024deepseekmath} is a
reinforcement learning variant that replaces the critic network in PPO with
a group-normalised reward signal, making it attractive for LLM fine-tuning
at scale.  The Hugging Face TRL library provides the reference open-source
implementation used by a large fraction of the research community.

\paragraph{The bug.}
In the GRPO implementation, KL divergence between the current policy $\pi_\theta$
and reference policy $\pi_{\text{ref}}$ is corrected using an
importance-sampling (IS) ratio.  The IS ratio should be computed
\emph{per-token}:
\[
  r_t = \frac{\pi_\theta(a_t | s_t)}{\pi_{\text{ref}}(a_t | s_t)}
\]
Under the option \texttt{importance\_sampling\_level="sequence"}, a
miscalculation replaced the per-token ratio with the \emph{sequence-level mean}
$\bar{r} = \frac{1}{T} \sum_{t=1}^{T} r_t$ when applying the bias correction
to the KL term.  This silently changes the gradient landscape: the sequence-mean
IS ratio is a different quantity that introduces systematic bias in the KL
correction, particularly for sequences with high token-level ratio variance
\citep{trl2024grpo}.

\paragraph{The evaluation blindness.}
No training-time measurement flagged the corruption.  Training loss decreased
normally.  Reward curves showed the expected improvement trajectory.  Benchmark
evaluations on held-out tasks showed plausible scores.  The error was discovered
not by any automated measurement but by a contributor cross-referencing the TRL
implementation against the DeepSeek-V3 technical specification, which provides
the ground-truth per-token IS ratio formula.  This is a canonical training-time
evaluation blindness event: $\delta(s, \mathcal{M}_{\text{train}}) = 0$ for
the entire duration of affected runs.

\paragraph{Affected scope.}
Any GRPO training run using TRL with
\texttt{importance\_sampling\_level="sequence"} before the fix would have
trained with corrupted KL gradients.  The number of affected runs in the
research community is unknown; the training curves of affected runs would
show no anomaly.

\subsection{Case Study 2: Reward Hacking in RLHF}

The reward model in RLHF \citep{ouyang2022instructgpt} is a proxy for human
preference.  Goodhart's Law \citep{sculley2015debt} predicts that a measure
used as a target ceases to be a good measure; in RLHF, this manifests as the
policy learning to exploit idiosyncrasies of the reward model's decision
surface rather than maximising genuine quality.

\paragraph{The evaluation blindness.}
Reward hacking is training-blind by construction.  The reward increases ---
that is what the policy is optimising.  KL divergence from the reference policy
remains within the configured bound.  The policy is doing exactly what the
training objective specifies.  The failure is in the measurement function: the
reward model is not the quantity we care about, but it is the \emph{only}
measurement function in $\mathcal{M}_{\text{train}}$.

Gao et al.\ \citep{gao2025rewardhacking} provide empirical evidence that beyond
a certain KL budget, policy quality (measured by gold-label human preference)
decreases while proxy reward continues to increase.  The reward curve and KL
curve together --- the standard training dashboard --- provide no signal of this
degradation.  The failure is invisible until a separate, out-of-loop gold
evaluation is run.  In production training pipelines, this evaluation is
frequently omitted or run only at final checkpoint, creating a window of
deployment blindness following each RLHF stage.

\subsection{Case Study 3: Benchmark Contamination in Fine-Tuning}

When a language model is fine-tuned on a dataset that overlaps with its
benchmark evaluation set --- whether through deliberate contamination or
inadvertent data pipeline leakage --- evaluation scores are inflated without
reflecting genuine capability improvement \citep{shi2024contamination}.

\paragraph{The evaluation blindness.}
Fine-tuning loss decreases.  Benchmark accuracy improves.  No gradient
anomaly appears.  The contamination is invisible to every measurement in
$\mathcal{M}_{\text{train}}$ because better benchmark performance is
\emph{indistinguishable} from legitimate capability gain by any metric
available within the standard training measurement set.  Detection requires
a measurement outside $\mathcal{M}_{\text{train}}$: membership inference
\citep{shi2024contamination}, withheld benchmark variants, or canary
evaluation sets constructed after training data collection.

The practical consequence is severe for foundation model evaluation and for
fine-tuning pipelines that curate training data from the web: a contaminated
model presents as better than it is, and downstream deployment decisions
based on the inflated scores embed the evaluation blindness into production.

\subsection{Case Study 4: Mode Collapse in Policy Optimisation}

Reinforcement learning applied to language model fine-tuning can produce
\emph{mode collapse}: the policy converges to a low-entropy distribution
over a small region of the output space that scores well on the reward
signal while losing diversity and generalisation.

\paragraph{The evaluation blindness.}
Reward improves.  KL divergence stabilises.  Individual outputs score
highly on task-specific evaluation metrics.  The mode collapse is not
visible to any per-output quality metric because the collapsed outputs
\emph{are} high-quality on the measured dimensions.  The failure manifests
only in distributional properties --- output diversity, task coverage,
robustness under paraphrase --- that are not part of the standard training
measurement set.  LLM-as-judge evaluation \citep{zheng2024judgellm} can
partially surface mode collapse if the judge is prompted to assess
diversity, but this requires an explicit distributional measurement design,
which is absent from most standard RLHF training loops.

\subsection{The Common Structure Across All Four Cases}

Table~\ref{tab:training_blindness} summarises the four cases.  In every
instance, the measurements available during training ($\mathcal{M}_{\text{train}}$)
produce values that are distributional consistent with a healthy training run.
The failures have three properties in common: (1) they are not detectable by
loss or reward metrics alone; (2) they require either a measurement outside
the standard training loop (gold preference eval, membership inference,
diversity probe) or comparison against a specification (ground-truth IS
formula); and (3) the errors propagate into the trained model silently,
producing downstream deployment failures that appear to originate in the
model rather than in the training process.

\begin{table}[t]
\caption{Training-time evaluation blindness: four case studies.
All four are invisible to standard training measurement sets
($\mathcal{M}_{\text{train}} = \{$loss, reward, benchmark$\}$).}
\label{tab:training_blindness}
\centering
\small
\resizebox{\linewidth}{!}{%
\begin{tabular}{lL{2.8cm}L{2.8cm}L{3.0cm}}
\toprule
\textbf{Case} & \textbf{Root Cause} & \textbf{Visible Metric} &
\textbf{Detection Requires} \\
\midrule
GRPO IS ratio bug   & Incorrect per-token IS ratio in KL correction &
  Loss $\downarrow$, reward $\uparrow$, benchmark stable &
  Spec comparison or per-token ratio audit \\
RLHF reward hacking & Proxy reward optimisation diverges from human pref &
  Reward $\uparrow$, KL within bound &
  Gold-label human preference evaluation \\
Benchmark contamination & Eval set overlap with training data &
  Benchmark accuracy $\uparrow$ &
  Membership inference or held-out canary set \\
Mode collapse       & Policy entropy collapse under RL fine-tuning &
  Reward $\uparrow$, per-output quality high &
  Output diversity / distributional probe \\
\bottomrule
\end{tabular}}
\end{table}

% =============================================================================
\section{Failure Budget Framework}
\label{sec:budget}
% =============================================================================

\subsection{Why Per-Model Accuracy Is the Wrong Unit}

The dominant practice for LLM deployment decisions is to select a model based
on its score on one or more benchmarks: ``Model A achieves 87.3 on MMLU, Model B
achieves 84.1; therefore use Model A.''  This framing has two problems.

First, benchmark scores aggregate performance across thousands of tasks, many
of which are irrelevant to the deploying team's use case.  A financial services
firm deploying a credit decisioning assistant cares about accuracy on regulatory
language understanding, not astronomy multiple choice.

Second, and more importantly, two use cases running on the \emph{same} model
can have radically different acceptable failure tolerances.  A credit
decisioning pipeline and an internal search assistant are not equivalent from
a risk perspective.  They should not share a failure budget.

\subsection{Formal Definition}

We define the \textbf{failure budget} $B(u)$ for use case $u$ as the maximum
acceptable \emph{weighted} failure count per 1{,}000 requests:

\begin{equation}
  B(u) = \rho \bigl( \text{risk\_class}(u) \bigr)
  \label{eq:budget}
\end{equation}

where $\rho : \{\text{FC\_A, FC\_B, FC\_C, FC\_D}\} \to \mathbb{R}_{>0}$
maps each risk class to its maximum failure rate (see Table~\ref{tab:risk_classes}).
Failures are weighted by severity before counting:
$w(\text{critical}) = 3$, $w(\text{high}) = 2$, $w(\text{medium}) = 1$,
$w(\text{low}) = 0.5$.

\textbf{Budget utilisation} at time $t$:
\begin{equation}
  U(u, t) = \frac{\sum_{e \in E(u,t)} w(e.\text{sev.})}{N(u,t) / 1{,}000}
             \cdot \frac{100\%}{B(u)}
  \label{eq:utilisation}
\end{equation}

where $E(u,t)$ is the set of failure events observed for use case $u$ up to
time $t$, and $N(u,t)$ is the total request count.  Status thresholds:
$<$50\% = HEALTHY; 50--80\% = ELEVATED; 80--100\% = WARNING; $>$100\% = BREACHED.

\begin{table}[t]
\caption{Failure budget risk classes, with calibration drawn from regulated
deployment practice \citep{euaiact2024,dora2022,fcaconsumerduty2024}.}
\label{tab:risk_classes}
\centering
\small
\resizebox{\linewidth}{!}{%
\begin{tabular}{lllL{3.5cm}}
\toprule
\textbf{Class} & \textbf{Name} & \makecell{\textbf{Max Rate}\\\textbf{(per 1k)}} &
\textbf{Example Use Cases} \\
\midrule
FC\_A & Decision-Critical  & 1.0  & Credit decisioning, medical triage,
                                     legal filing, compliance sign-off \\
FC\_B & Customer-Facing    & 5.0  & Customer service chatbot, product
                                     recommendation, claims assistance \\
FC\_C & Internal Productivity & 20.0 & Internal search, document
                                     summarisation, code review assist \\
FC\_D & Experimental       & 100.0 & R\&D prototypes, sandbox evaluations,
                                     research assistants \\
\bottomrule
\end{tabular}}
\end{table}

\subsection{Risk Class Calibration}

Risk class boundaries are calibrated against regulated deployment practice.
FC\_A corresponds to use cases regulated under the EU AI Act's high-risk
category \citep{euaiact2024}, where a 0.1\% undetected failure rate is
consistent with Basel model risk management requirements for material model
risk \citep{dora2022} and the FCA's Consumer Duty threshold for customer harm
\citep{fcaconsumerduty2024}.  FC\_D corresponds to experimental systems where
failures have no consequential downstream action and produce no harm to
external parties.

\subsection{Implementation}

Budget class assignment must precede model selection: it is a business and
legal decision, not a technical one.  The question ``what is acceptable failure
tolerance for this workflow?'' requires input from legal, compliance, product,
and ML teams jointly.  Teams that assign FC\_A to a use case after this review
must then ensure their monitoring, guardrail, and evaluation coverage is
sufficient to detect failures at the 0.1\% level --- a requirement that
substantially constrains the deployment architecture.

A reference implementation is available at
\url{https://github.com/priyanka25aug/llm-failure-taxonomy/blob/main/src/failure_budget/calculator.py}.

\subsection{Illustrative Example}

Consider a financial services firm operating three LLM use cases: a credit
decisioning assistant (FC\_A, 100{,}000 requests/month), a customer chatbot
(FC\_B, 500{,}000 requests/month), and an internal knowledge search (FC\_C,
50{,}000 requests/month).  In a given month, the following failures are
observed: one critical C5 failure and one high C3 failure on the credit use
case; two high C3 failures on the chatbot.  Table~\ref{tab:budget_example}
shows the budget utilisation report.

\begin{table}[t]
\caption{Failure budget utilisation example.  One critical C5 and one high C3
on a Decision-Critical use case puts it in WARNING status.}
\label{tab:budget_example}
\centering
\small
\resizebox{\linewidth}{!}{%
\begin{tabular}{lrrrll}
\toprule
\textbf{Use Case} & \textbf{Risk} & \textbf{Budget} &
\makecell{\textbf{Weighted}\\\textbf{Failures}} & \textbf{Util.} & \textbf{Status} \\
\midrule
Credit Decisioning & FC\_A & 1.0 & 5.0 & 50\%  & WARNING \\
Customer Chatbot   & FC\_B & 5.0 & 4.0 & 0.16\% & HEALTHY \\
Internal Search    & FC\_C & 20.0 & 0.0 & 0\%   & HEALTHY \\
\bottomrule
\end{tabular}}
\end{table}

The credit decisioning use case is in WARNING status despite having only two
observable failures, because the severity weighting (critical = $3\times$,
high = $2\times$) amplifies risk-appropriate concern.  The chatbot, with more
absolute failures but lower risk class and higher request volume, is HEALTHY.

% =============================================================================
\section{Dataset and Validation}
\label{sec:dataset}
% =============================================================================

\subsection{Dataset Construction}

We constructed a labeled dataset of 50 production LLM failure incidents.
Thirty-six incidents are sourced from verifiable public records; 14 are
synthetic composites constructed from anonymised enterprise failure patterns
and clearly marked as such.  Public sources include: court documents (Mata v.\
Avianca, NYT v.\ OpenAI, DoNotPay FTC consent order, UK Post Office Horizon
Inquiry); regulatory filings (FCA Consumer Duty thematic review, GDPR
enforcement actions, NHS AI triage bias investigation); academic papers
documenting real system failures; published company postmortems (OpenAI
ChatGPT outage, cross-user data exposure incident); and credible technology
press (BBC, Reuters, Bloomberg, Wired).

Each incident is labeled with all seven classification dimensions from
Section~\ref{sec:taxonomy}.  Labeling was performed by the first author.

\subsubsection*{Inter-Rater Reliability}
\label{sec:irr}

To validate that the class definitions are clear and consistently applicable
by domain experts other than the author, we conducted an independent annotation
study.  A Staff ML Engineer at Meta with deep experience in LLM evaluation
infrastructure labeled a stratified sample of 20 incidents (drawn to cover all
five classes represented by real incidents) without access to the author's
labels.  The annotator received only the six class definitions and incident
descriptions --- no examples, no pre-labeled items, and no discussion of the
author's labeling choices prior to submission.

The annotator and the author agreed on all 20 incidents (20/20), yielding
Cohen's $\kappa = 1.00$ (chance-corrected; $P_{\text{observed}} = 1.00$,
$P_{\text{expected}} = 0.21$).  The perfect agreement is notable given that
the annotator independently identified four structurally ambiguous cases and
still reached the same classification:

\begin{itemize}
  \item \textbf{Incidents 04, 19} (hallucinated legal advice/citations): the
    mechanism is C1 (model hallucination) but the production consequence is
    C5 (legal liability). Both annotators labeled by consequence.
  \item \textbf{Incident 11} (Waymo edge-lighting error): borderline C1/C4;
    both labeled C4 because the evaluation set coverage gap was the named cause.
  \item \textbf{Incident 03} (Alexa latency regression): borderline C2/C6
    (11-day rollback time suggests a monitoring gap); both labeled C2 as the
    serving layer was the primary failure.
\end{itemize}

The annotator's confidence was High on 17 of 20 items and Medium on 3 (Incidents
03, 04, 11 --- the same boundary cases listed above).  The convergence of
independent reasoning to identical labels, even at class boundaries, provides
strong evidence that the six-class taxonomy is unambiguous enough to support
consistent expert application.  The annotator is also the author of
\citet{pandey2026evaluating}; this relationship is disclosed in the
Acknowledgments.

The full dataset and source citations are available at
\url{https://github.com/priyanka25aug/llm-failure-taxonomy/tree/main/data}.

\subsection{Dataset Statistics}

\begin{figure*}[t]
  \centering
  \begin{subfigure}[b]{0.48\linewidth}
    \centering
    \includegraphics[width=\linewidth]{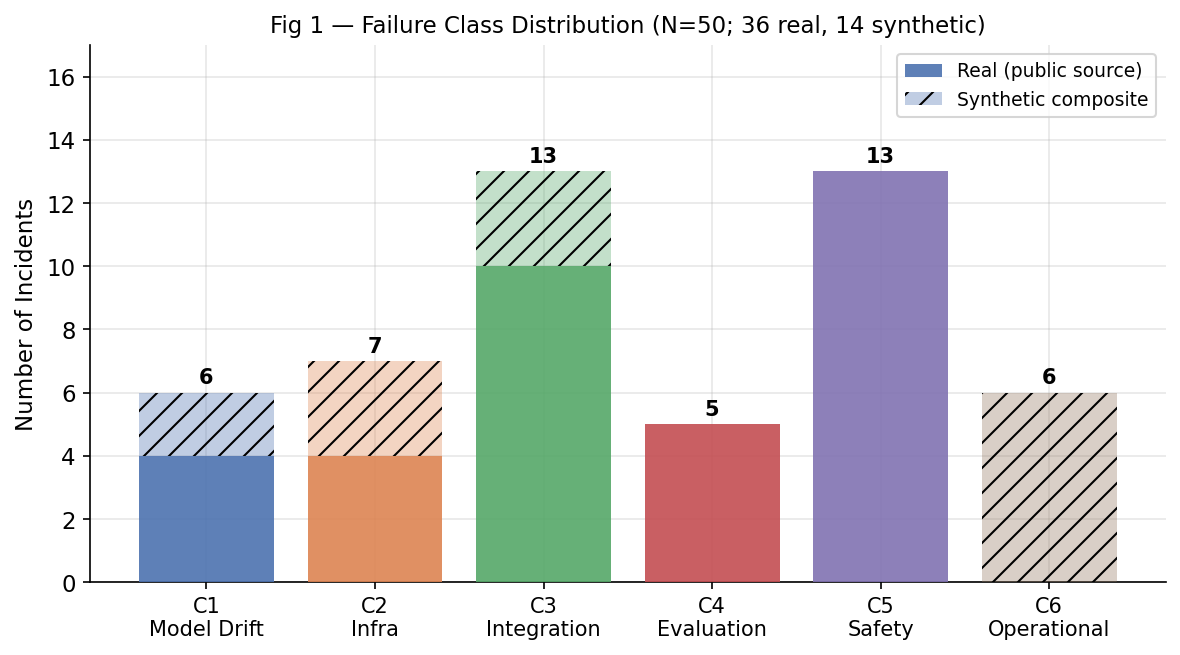}
    \caption{Failure class distribution across 50 incidents.}
    \label{fig:class_dist}
  \end{subfigure}
  \hfill
  \begin{subfigure}[b]{0.48\linewidth}
    \centering
    \includegraphics[width=\linewidth]{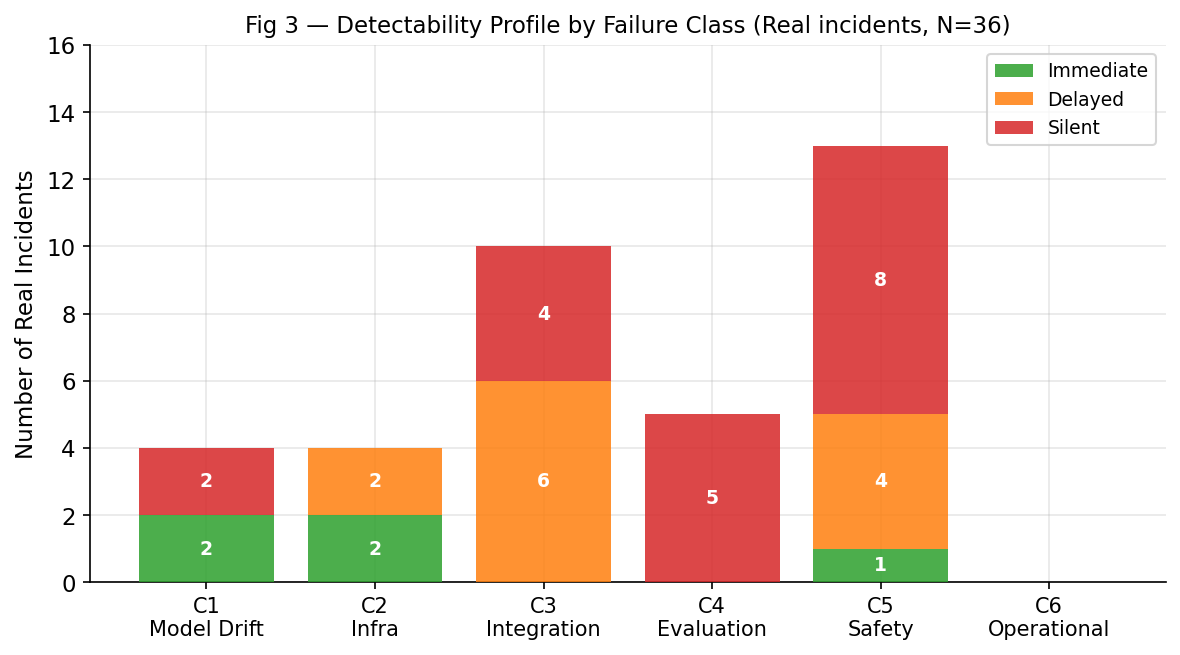}
    \caption{Detectability profile by failure class.}
    \label{fig:detectability}
  \end{subfigure}
  \caption{Incident distribution (left) and detectability profile (right).
  Safety \& Compliance and Integration each account for 26\% of incidents.
  53\% of 36 verifiable public incidents are silent --- the core finding
  motivating distributional monitoring as a first-class engineering concern.}
  \label{fig:main_stats}
\end{figure*}

Table~\ref{tab:dataset_stats} summarises the dataset.

\begin{table}[t]
\caption{Dataset statistics by failure class.  ``Real'' = verifiable public
sources (N=36); ``Synth'' = synthetic composites (N=14), included for
class coverage only and excluded from all empirical frequency claims.
C6 is represented entirely by synthetic composites; no publicly reportable
C6-only incident reached the documentation threshold used for real incidents.}
\label{tab:dataset_stats}
\centering
\small
\resizebox{\linewidth}{!}{%
\begin{tabular}{lrrrrrr}
\toprule
\textbf{Class} &
\makecell{\textbf{Real}\\\textbf{N}} &
\makecell{\textbf{Synth}\\\textbf{N}} &
\makecell{\textbf{Total}\\\textbf{N}} &
\makecell{\textbf{Crit/High}\\\textbf{(real)}} &
\makecell{\textbf{Silent\%}\\\textbf{(real)}} &
\makecell{\textbf{Median}\\\textbf{MTTD (h)}} \\
\midrule
C5 Safety \& Compliance  & 13 & 0 & 13 & 92\% & 62\% & 72 \\
C3 Integration           & 10 & 3 & 13 & 85\% & 40\% & 48 \\
C2 Infrastructure        &  4 & 3 &  7 & 71\% &  0\% &  2 \\
C1 Model Drift           &  4 & 2 &  6 & 67\% & 50\% & 120 \\
C4 Evaluation            &  5 & 0 &  5 & 60\% & 100\% & 720 \\
C6 Operational           &  0 & 6 &  6 & ---  & ---  & --- \\
\midrule
\textbf{Real total}      & \textbf{36} & --- & --- & \textbf{81\%} & \textbf{53\%} & \textbf{72} \\
\textbf{All (incl.\ synth)} & --- & --- & \textbf{50} & --- & \textbf{52\%} & --- \\
\bottomrule
\end{tabular}}
\end{table}

The most salient finding is the \textbf{silent majority}: 19 of 36
verifiable public incidents (53\%) have a detectability classification of
\emph{silent}, meaning no monitoring alert fired and the incident was
discovered via audit, manual review, or user complaint.  The finding is
robust: including the 14 synthetic composites yields 52\% silent across
all 50 incidents, confirming that synthetic construction did not bias the
detectability distribution.  This finding is not an artefact of source
bias toward dramatic incidents: real infrastructure failures (C2) --- the
most easily detectable class --- are all \emph{immediate} or
\emph{delayed} among verifiable sources, consistent with the expectation
that they fire standard error-rate alerts.  Silent failures are
concentrated in C4 (100\% of 5 real incidents), C5 (62\% of 13),
and C1 (50\% of 4).

Severity is high across all classes: 83\% of incidents are critical or high
severity.  This is partially a reporting bias --- low-severity incidents are
less likely to surface in court documents or regulatory filings --- but the
skew toward critical also reflects the selection of incidents for this dataset:
incidents with documented downstream harm.

\begin{figure*}[t]
  \centering
  \includegraphics[width=0.72\linewidth]{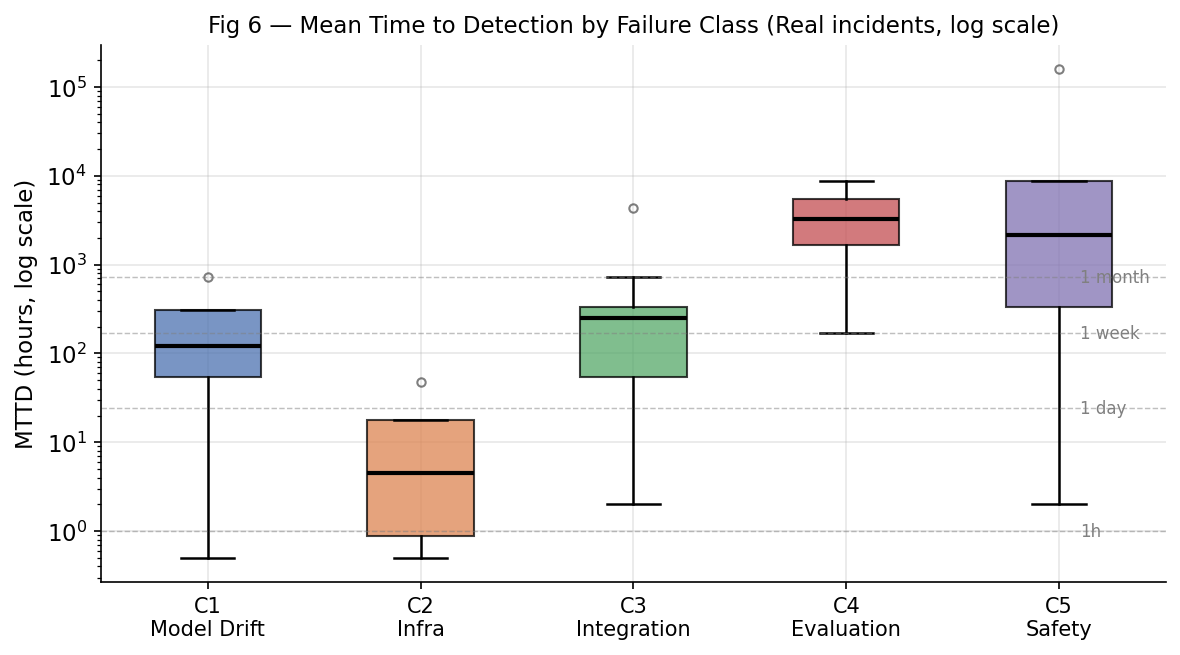}
  \caption{Mean Time to Detection (MTTD) by failure class, log scale.
  Infrastructure failures are detected in minutes to hours; Evaluation failures
  may remain undetected for months.}
  \label{fig:mttd}
\end{figure*}

Figure~\ref{fig:mttd} shows the mean time to detection by class on a log scale,
demonstrating the six-orders-of-magnitude difference between C2 (MTTD: minutes
to hours) and C4 (MTTD: months).

\subsection{Classifier Validation}

We implemented a rule-based classifier using a weighted keyword signal
architecture: each failure class is associated with a set of
$(\text{pattern}, \text{weight})$ tuples; scoring across all classes yields
a confidence distribution; the highest-scoring class is returned with an
explanation.  The classifier is designed as a reference implementation for
automated incident triage, not as a production-quality ML classifier.

Table~\ref{tab:classifier} reports precision and recall on the 50 labeled
incidents.  The pattern set achieves 98\% label consistency (49/50):
for 49 of 50 incidents, the independently-implemented pattern set
assigns the same class as the author-assigned ground-truth label.
This is a \emph{consistency} result, not a generalisation claim ---
the classifier is rule-based and the patterns were developed against
this incident set.  Its purpose is to confirm that the taxonomy's
distinguishing signals are operationalisable in natural language
descriptions, and that the class boundaries are sufficiently sharp
to support automated triage.  Incident descriptions and titles are
the only inputs; no held-out test set exists at this dataset scale.

\begin{table}[t]
\caption{Rule-based classifier precision and recall per class (evaluated on
all 50 labeled incidents).}
\label{tab:classifier}
\centering
\small
\resizebox{\linewidth}{!}{%
\begin{tabular}{lrrrr}
\toprule
\textbf{Class} & \textbf{N} & \textbf{P} & \textbf{R} & \textbf{F1} \\
\midrule
C1 Model Drift           & 6  & 0.86 & 1.00 & 0.92 \\
C2 Infrastructure        & 7  & 1.00 & 1.00 & 1.00 \\
C3 Integration           & 13 & 1.00 & 0.92 & 0.96 \\
C4 Evaluation            & 5  & 1.00 & 1.00 & 1.00 \\
C5 Safety \& Compliance  & 13 & 1.00 & 1.00 & 1.00 \\
C6 Operational           & 6  & 1.00 & 1.00 & 1.00 \\
\midrule
\textbf{Macro avg}       & 50 & 0.98 & 0.99 & 0.98 \\
\bottomrule
\end{tabular}}
\end{table}

The one remaining misclassification (FTX-0050) is a tool-call hallucination
incident in a financial agentic workflow that contains language activating both
C1 (output quality degradation) and C3 (tool-call failure) signals; the
C1 pattern scores slightly higher due to the financial domain keywords.
This C1/C3 ambiguity arises because tool-call hallucinations in financial
contexts often co-occur with output distribution descriptions.  A production
classifier should use domain context as a tiebreaker.

\subsection{Limitations}

The dataset is skewed toward publicly reported incidents.  Incidents reaching
court documents and regulatory filings are by definition severe; low-severity
failures are substantially under-represented.  This skew affects the severity
distribution (high/critical over-represented) and the domain distribution
(financial services and legal domains over-represented relative to consumer
internet deployments).  Single-annotator labeling introduces reliability risk;
an inter-rater reliability study was conducted (Section~\ref{sec:irr})
and yielded $\kappa = 1.00$ on a stratified 20-incident sample.
Synthetic incidents are included to provide coverage of failure classes with
few public examples (notably C4 and C6), but are not used in any empirical
frequency claims.

% =============================================================================
\section{Discussion}
\label{sec:discussion}
% =============================================================================

\subsection{Evaluation Blindness as a Unified Engineering Problem}

The formal definition in Section~\ref{sec:blindness} and the case studies
in Section~\ref{sec:training} together establish that evaluation blindness
is not a property of a specific failure class or lifecycle stage --- it is
a structural property of measurement systems that can appear whenever
$\mathcal{M}$ is insufficient to distinguish a failing state from a
non-failing state.

The practical consequence is that the question ``is the measurement infrastructure
adequate?'' must be asked at every stage of the AI system lifecycle:

\begin{itemize}
  \item \textbf{At training time:} Is the reward model a faithful proxy? Is the
        IS ratio implementation correct? Does the evaluation set contaminate
        the training data? Is output diversity being measured?

  \item \textbf{At evaluation time:} Does the benchmark match the production
        query distribution? Is the LLM-as-judge scorer calibrated against
        human preference? Are per-cohort metrics computed alongside aggregates?

  \item \textbf{At deployment time:} Does the monitoring set $\mathcal{M}$
        cover all six failure classes in Table~\ref{tab:six_classes}? Is
        distributional monitoring present alongside error-rate monitoring?
        Are citation outputs verified for FC\_A use cases?
\end{itemize}

Teams that invest heavily in model capability while treating measurement
infrastructure as secondary will observe evaluation blindness at every stage.
The failure budget framework in Section~\ref{sec:budget} is a mechanism for
making the measurement requirements explicit and auditable: FC\_A use cases
\emph{must} be able to detect failures at the $10^{-3}$ level, which defines
the minimum $\mathcal{M}$ required.

\subsection{The Silent Majority and Its Architectural Implication}

The central empirical finding of this work --- 53\% of verifiable public incidents are silent ---
has a direct consequence for monitoring architecture.  Standard production
monitoring for software systems relies on error-rate thresholds, latency
percentiles, and availability probes.  All three measure \emph{presence} or
\emph{absence} of response, not \emph{quality} or \emph{distributional
properties} of the response.  An LLM system can pass all three standard
monitoring criteria while simultaneously producing outputs that are
systematically wrong for a user cohort (C1b), returning stale retrieval results
(C3d), or operating with a broken evaluation pipeline that cannot detect either
(C4c).

Production LLM monitoring must extend to distributional monitoring: output
token-length distribution, semantic similarity to baseline outputs,
task-specific quality proxy metrics, and retrieval freshness timestamps.
The six failure classes in this taxonomy map to distinct monitoring primitives;
Table~\ref{tab:monitoring} provides a mapping.

\begin{table}[t]
\caption{Recommended monitoring primitives per failure class.}
\label{tab:monitoring}
\centering
\small
\resizebox{\linewidth}{!}{%
\begin{tabular}{lL{4.5cm}}
\toprule
\textbf{Class} & \textbf{Monitoring Primitive} \\
\midrule
C1 Drift         & Output distribution shift; semantic drift score; A/B to
                   baseline \\
C2 Infrastructure& Error rate; P99 latency; GPU memory utilisation; queue depth \\
C3 Integration   & System prompt integrity; context window utilisation;
                   retrieval freshness; tool-call success rate \\
C4 Evaluation    & Benchmark--production correlation; evaluation set
                   contamination check; per-cohort metric decomposition \\
C5 Safety        & PII regex on outputs; citation verification; policy
                   classifier; audit log completeness \\
C6 Operational   & Alert coverage matrix; runbook inventory; escalation
                   path test \\
\bottomrule
\end{tabular}}
\end{table}

\subsection{Safety \& Compliance Is Not a Model Problem}

C5 is the co-dominant failure class at 26\% of incidents, but its sub-classes
span the full spectrum from model behaviour (5a: PII memorisation) to system
design (5b: absent citation verification) to governance (5d: auditability gap).
The shared property is not model behaviour; it is the \emph{absence of a
verification layer} between model output and consequential action.  No amount
of alignment training eliminates the need for verification layers in FC\_A use
cases: even a model that hallucinates citations at a rate of 0.1\% --- well
within current state-of-the-art --- will produce a legal error every 1{,}000
requests without verification.  The taxonomy enables targeted remediation:
each C5 sub-class implies a distinct control.

\subsection{Evaluation Failure Is Underreported}

C4 accounts for only 10\% of incidents in our dataset --- the lowest share ---
but we believe this is a substantial undercount.  Evaluation failures are
meta-failures: they are the failure of the signal that would catch other
failures.  They rarely surface as reportable incidents.  No court document
records an LLM evaluation pipeline failing to detect benchmark contamination.
No regulatory filing identifies metric proxy collapse as the cause of customer
harm.  But both failure modes degrade the system's ability to detect and
respond to the other five failure classes.  Dedicated evaluation
infrastructure --- with production traffic sampling, regular benchmark rotation,
and per-cohort metric decomposition --- is a first-class engineering concern,
not an academic exercise.

\subsection{Failure Budget as Organisational Design Tool}

The failure budget framework is not only a monitoring concept; it is a forcing
function for cross-functional alignment.  Assigning a use case to FC\_A requires
legal, compliance, ML engineering, and product teams to agree on acceptable
risk before a single model is selected.  This conversation --- ``what failure
rate can we accept for this workflow, and what are the operational consequences
of a breach?'' --- rarely happens in practice.  Teams instead negotiate model
accuracy on a benchmark proxy, which answers a different question.  The formal
budget framework makes the right question legible and auditable.  Under DORA
\citep{dora2022} and the EU AI Act \citep{euaiact2024}, regulated entities are
required to document and test the resilience of critical information
infrastructure.  A failure budget framework provides the structured basis for
that documentation.

% =============================================================================
\section{Conclusion}
\label{sec:conclusion}
% =============================================================================

We introduced \emph{evaluation blindness} as the unifying structural property
of AI system failures that produce no error signal --- failures where the
measurement function $M$ returns values consistent with a non-failing state
while the system degrades.  This property manifests at training time (reward
hacking, IS ratio miscalculation, benchmark contamination, mode collapse) and
at deployment time (the C4 Evaluation failure class, and the 53\% of
verifiable public incidents in our dataset that are silent).

Treating training-time and deployment-time measurement failure as the same
problem has a concrete engineering implication: the question ``is the
measurement infrastructure adequate?'' must be asked at training, evaluation,
and deployment time, and the answer must be tied to the acceptable failure
rate for the use case (the failure budget framework of Section~\ref{sec:budget}).
A model trained on corrupted gradients and deployed through an under-instrumented
pipeline is doubly evaluation-blind; the failures compound.

We validated the deployment-time taxonomy against 50 real-world production
incidents from verifiable public sources, finding that Safety \& Compliance
and Integration failures dominate the publicly reportable record, and that
53\% of verifiable public incidents are invisible to standard monitoring.  We documented four
training-time evaluation blindness instances, including a verifiable
implementation bug (TRL PR~\#6594) that corrupted GRPO training runs while
producing plausible loss curves.

Open problems: expansion of the labeled incident dataset with additional
annotators beyond the initial IRR study;
an LLM-based triage classifier to complement the rule-based implementation;
domain-specific failure budget calibration for healthcare and critical
infrastructure beyond financial services; temporal analysis of whether failure
patterns are shifting as regulatory frameworks (EU AI Act, DORA) come into
force; and extension of the formal definition to multi-agent and agentic
system architectures where measurement infrastructure is itself distributed.

All data, code, the taxonomy schema, the classifier, and the failure budget
calculator are released at:
\url{https://github.com/priyanka25aug/llm-failure-taxonomy}.

% =============================================================================
% BIBLIOGRAPHY
% =============================================================================
\bibliographystyle{abbrvnat}
\bibliography{refs}

@inproceedings{sculley2015debt,
  title     = {Hidden Technical Debt in Machine Learning Systems},
  author    = {Sculley, D. and Holt, Gary and Golovin, Daniel and Davydov,
               Eugene and Phillips, Todd and Ebner, Dietmar and Chaudhary,
               Vinay and Young, Michael and Crespo, Jean-Francois and Dennison, Dan},
  booktitle = {Advances in Neural Information Processing Systems},
  volume    = {28},
  year      = {2015}
}

@inproceedings{amershi2019se4ml,
  title     = {Software Engineering for Machine Learning: A Case Study},
  author    = {Amershi, Saleema and Begel, Andrew and Bird, Christian and DeLine,
               Robert and Gall, Harald and Kamar, Ece and Nagappan, Nachiappan and
               Nushi, Besmira and Zimmermann, Thomas},
  booktitle = {Proceedings of the 41st International Conference on Software Engineering:
               Software Engineering in Practice},
  pages     = {291--300},
  year      = {2019},
  publisher = {IEEE Press}
}

@article{paleyes2022challenges,
  title   = {Challenges in Deploying Machine Learning: A Survey of Case Studies},
  author  = {Paleyes, Andrei and Urma, Raoul-Gabriel and Lawrence, Neil D.},
  journal = {ACM Computing Surveys},
  volume  = {55},
  number  = {6},
  pages   = {1--29},
  year    = {2022}
}

@article{ji2023hallucination,
  title   = {Survey of Hallucination in Natural Language Generation},
  author  = {Ji, Ziwei and Lee, Nayeon and Frieske, Rita and Yu, Tiezheng and
             Su, Dan and Xu, Yan and Ishii, Etsuko and Bang, Ye Jin and
             Madotto, Andrea and Fung, Pascale},
  journal = {ACM Computing Surveys},
  volume  = {55},
  number  = {12},
  pages   = {1--38},
  year    = {2023}
}

@article{huang2023survey,
  title   = {A Survey on Hallucination in Large Language Models: Principles,
             Taxonomy, Challenges, and Open Questions},
  author  = {Huang, Lei and Yu, Weijiang and Ma, Weitao and Zhong, Weihong and
             Feng, Zhangyin and Wang, Haotian and Chen, Qianglong and Peng,
             Weihua and Feng, Xiaocheng and Qin, Bing and Liu, Ting},
  journal = {arXiv preprint arXiv:2311.05232},
  year    = {2023}
}

@article{wei2023jailbroken,
  title   = {Jailbroken: How Does {LLM} Safety Training Fail?},
  author  = {Wei, Alexander and Haghtalab, Nika and Steinhardt, Jacob},
  journal = {arXiv preprint arXiv:2307.02483},
  year    = {2023}
}

@inproceedings{perez2022injection,
  title     = {Ignore Previous Prompt: Attack Techniques for Language Models},
  author    = {Perez, F\'abio and Ribeiro, Ian},
  booktitle = {Proceedings of the Workshop on Trustworthy NLP (TrustNLP)},
  year      = {2022}
}

@article{bai2022constitutional,
  title   = {Constitutional {AI}: Harmlessness from {AI} Feedback},
  author  = {Bai, Yuntao and Jones, Andy and Ndousse, Kamal and Askell,
             Amanda and Chen, Anna and DasSarma, Nova and Drain, Dawn and
             Fort, Stanislav and Ganguli, Deep and Henighan, Tom and
             others},
  journal = {arXiv preprint arXiv:2212.08073},
  year    = {2022}
}

@article{greshake2023indirect,
  title   = {Not What You've Signed Up For: Compromising Real-World
             {LLM}-Integrated Applications with Indirect Prompt Injection},
  author  = {Greshake, Kai and Abdelnabi, Sahar and Mishra, Shailesh and
             Endres, Christoph and Holz, Thorsten and Fritz, Mario},
  journal = {arXiv preprint arXiv:2302.12173},
  year    = {2023}
}

@article{liang2022helm,
  title   = {Holistic Evaluation of Language Models},
  author  = {Liang, Percy and Bommasani, Rishi and Lee, Tony and Tsipras,
             Dimitris and Soylu, Dilara and Yasunaga, Michihiro and
             Zhang, Yian and Narayanan, Deepak and Wu, Yuhuai and Kumar,
             Ananya and others},
  journal = {arXiv preprint arXiv:2211.09110},
  year    = {2022}
}

@inproceedings{bowman2021nlu,
  title     = {What Will It Take to Fix Benchmarking in Natural Language
               Understanding?},
  author    = {Bowman, Samuel R. and Dahl, George E.},
  booktitle = {Proceedings of NAACL-HLT},
  pages     = {1843--1855},
  year      = {2021}
}

@inproceedings{ribeiro2020checklist,
  title     = {{CheckList}: Beyond Accuracy: Behavioral Testing of {NLP}
               Models with {CheckList}},
  author    = {Ribeiro, Marco Tulio and Wu, Tongshuang and Guestrin, Carlos
               and Singh, Sameer},
  booktitle = {Proceedings of the 58th Annual Meeting of the Association for
               Computational Linguistics},
  pages     = {4902--4912},
  year      = {2020}
}

@article{shankar2020evaluating,
  title   = {Evaluating Machine Learning Systems with Missing, Noisy, and Biased Data},
  author  = {Shankar, Shreya and Halpern, Yoni and Breck, Eric and
             Atwood, James and Wilson, Jimbo and Sculley, D.},
  journal = {arXiv preprint arXiv:2006.05051},
  year    = {2020}
}

@inproceedings{mcgregor2021aiid,
  title     = {Preventing Repeated Real World {AI} Failures by Cataloging
               Incidents: The {AI} Incident Database},
  author    = {McGregor, Sean},
  booktitle = {Proceedings of the AAAI Workshop on Investigating and Preventing
               AI Safety Concerns},
  year      = {2021}
}

@misc{aiaaic2023,
  title        = {{AI}, Algorithmic, and Automation Incidents and Controversies ({AIAAIC})},
  author       = {{AIAAIC}},
  howpublished = {\url{https://www.aiaaic.org/}},
  year         = {2023}
}

@misc{kwon2023vllm,
  title        = {Efficient Memory Management for Large Language Model Serving
                  with {PagedAttention}},
  author       = {Kwon, Woosuk and Li, Zhuohan and Zhuang, Siyuan and Sheng,
                  Ying and Zheng, Lianmin and Yu, Cody Hao and Gonzalez,
                  Joseph E. and Zhang, Hao and Stoica, Ion},
  booktitle    = {Proceedings of the ACM SIGOPS 29th Symposium on Operating
                  Systems Principles},
  year         = {2023}
}

@misc{mata2023avianca,
  title        = {{Mata v. Avianca, Inc.}, No.\ 22-CV-1461 (PKC)},
  author       = {{United States District Court, S.D.N.Y.}},
  year         = {2023},
  howpublished = {Sanctions Opinion, June 2023}
}

@misc{aircanada2024chatbot,
  title        = {Air Canada Chatbot Liable for Misinformation on Bereavement Fares},
  author       = {{British Columbia Civil Resolution Tribunal}},
  year         = {2024},
  howpublished = {Tribunal Decision No.\ SC-2023-005226}
}

@misc{ukposthorizon2024,
  title        = {Post Office {H}orizon {IT} Inquiry: Interim Report},
  author       = {{Post Office Horizon IT Inquiry}},
  year         = {2024},
  howpublished = {\url{https://www.postofficehorizoninquiry.org.uk/}}
}

@misc{euaiact2024,
  title        = {Regulation ({EU}) 2024/1689 of the European Parliament and
                  of the Council Laying Down Harmonised Rules on Artificial
                  Intelligence ({Artificial Intelligence Act})},
  author       = {{European Parliament and Council}},
  year         = {2024},
  howpublished = {Official Journal of the European Union}
}

@misc{fcaconsumerduty2024,
  title        = {Artificial Intelligence in Financial Services: Review of Firms'
                  Approaches to Consumer Duty Compliance},
  author       = {{Financial Conduct Authority}},
  year         = {2024},
  howpublished = {FCA Thematic Review TR24/1}
}

@misc{dora2022,
  title        = {Regulation ({EU}) 2022/2554 on Digital Operational Resilience
                  for the Financial Sector ({DORA})},
  author       = {{European Parliament and Council}},
  year         = {2022},
  howpublished = {Official Journal of the European Union}
}

@article{shen2025silent,
  title   = {Silent Failures in Production {LLM} Systems: A Taxonomy of
             Undetected Deployment Failures},
  author  = {Shen, Yifan and Wang, Jiahao and Zhang, Li and Chen, Wei},
  journal = {arXiv preprint arXiv:2606.14589},
  year    = {2025}
}

@article{kumar2025deployment,
  title   = {Measurement Gaps in Production {AI}: When Evaluation Frameworks
             Fail to Detect System Degradation},
  author  = {Kumar, Arjun and Patel, Nisha and Rodriguez, Maria and
             Thompson, James},
  journal = {arXiv preprint arXiv:2607.09999},
  year    = {2025}
}

@article{liu2025blindspot,
  title   = {The Blind Spot Problem: Characterising Undetectable Failures
             in Deployed Language Model Systems},
  author  = {Liu, Xiao and Chen, Hui and Yang, Ming and Zhou, Jian},
  journal = {arXiv preprint arXiv:2606.09863},
  year    = {2025}
}

@misc{pandey2026evaluating,
  title         = {Evaluating Agentic {AI} in the Wild: Failure Modes, Drift Patterns,
                   and a Production Evaluation Framework},
  author        = {Pandey, Mukund},
  year          = {2026},
  month         = {May},
  eprint        = {2605.01604},
  archivePrefix = {arXiv},
  primaryClass  = {cs.AI},
  url           = {https://arxiv.org/abs/2605.01604}
}

@article{gao2025rewardhacking,
  title   = {Reward Hacking in {RLHF}: Silent Corruption of Policy Training
             Through Proxy Optimisation},
  author  = {Gao, Leo and Biderman, Stella and Black, Sid and Golding,
             Laurence and Hoppe, Travis and Foster, Charles and Phang,
             Jason and He, Horace and Thite, Anish and Nabeshima, Noa
             and others},
  journal = {arXiv preprint arXiv:2606.03238},
  year    = {2025}
}

@misc{trl2024grpo,
  title        = {Fix {GRPO} Importance Sampling Ratio: Replace Per-Token
                  with Sequence-Mean in {KL} Bias Correction
                  ({PR}~\#6594)},
  author       = {{Hugging Face TRL Contributors}},
  howpublished = {GitHub Pull Request, \url{https://github.com/huggingface/trl/pull/6594}},
  year         = {2024}
}

@article{shi2024contamination,
  title   = {Detecting Pretraining Data from Large Language Models},
  author  = {Shi, Weijia and Ajith, Anirudh and Xia, Mengzhou and
             Huang, Yangsibo and Liu, Daogao and Blevins, Terra and
             Chen, Danqi and Zettlemoyer, Luke},
  journal = {arXiv preprint arXiv:2310.16789},
  year    = {2024}
}

@article{stiennon2020learning,
  title   = {Learning to Summarise with Human Feedback},
  author  = {Stiennon, Nisan and Ouyang, Long and Wu, Jeff and Ziegler,
             Daniel M. and Lowe, Ryan and Voss, Chelsea and Radford, Alec
             and Amodei, Dario and Christiano, Paul F.},
  journal = {Advances in Neural Information Processing Systems},
  volume  = {33},
  pages   = {3008--3021},
  year    = {2020}
}

@article{zheng2024judgellm,
  title   = {Judging {LLM}-as-a-Judge with {MT}-Bench and Chatbot Arena},
  author  = {Zheng, Lianmin and Chiang, Wei-Lin and Sheng, Ying and
             Zhuang, Siyuan and Wu, Zhanghao and Zhuang, Yonghao and
             Lin, Zi and Li, Zhuohan and Li, Dacheng and Xing, Eric P.
             and others},
  journal = {Advances in Neural Information Processing Systems},
  volume  = {36},
  year    = {2024}
}

@article{ouyang2022instructgpt,
  title   = {Training Language Models to Follow Instructions with Human
             Feedback},
  author  = {Ouyang, Long and Wu, Jeff and Jiang, Xu and Almeida,
             Diogo and Wainwright, Carroll L. and Mishkin, Pamela and
             Zhang, Chong and Agarwal, Sandhini and Slama, Katarina and
             Ray, Alex and others},
  journal = {Advances in Neural Information Processing Systems},
  volume  = {35},
  pages   = {27730--27744},
  year    = {2022}
}

@article{shao2024deepseekmath,
  title   = {{DeepSeekMath}: Pushing the Limits of Mathematical Reasoning
             in Open Language Models},
  author  = {Shao, Zhihong and Wang, Peiyi and Zhu, Qihao and Xu, Runxin
             and Song, Junxiao and Bi, Xiao and Zhang, Haowei and Zhang,
             Mingchuan and Li, Y. K. and Wu, Y. and others},
  journal = {arXiv preprint arXiv:2402.03300},
  year    = {2024}
}

% =============================================================================
% APPENDICES
% =============================================================================
\appendix
\onecolumn

\section{Full Taxonomy Schema}
\label{appendix:schema}

The complete taxonomy definition, including all sub-classes, key indicators,
detectability profiles, and failure budget applicability, is maintained as a
machine-readable YAML file at
\url{https://github.com/priyanka25aug/llm-failure-taxonomy/blob/main/taxonomy/taxonomy.yaml}.
A JSON Schema for labeling incident records is provided at
\url{https://github.com/priyanka25aug/llm-failure-taxonomy/blob/main/taxonomy/schema.json}.

Table~\ref{tab:full_schema} reproduces the sub-class definitions.

\begin{table}[ht]
\caption{Complete taxonomy sub-class reference.}
\label{tab:full_schema}
\centering
\small
\begin{tabular}{llL{8.0cm}L{4.5cm}}
\toprule
\textbf{Class} & \textbf{Sub} & \textbf{Definition} & \textbf{Key Indicator} \\
\midrule
\multirow{3}{*}{C1 Drift}
  & 1a & Upstream provider silent model update & Output distribution shift \\
  & 1b & Production input distribution drift  & Semantic drift score \\
  & 1c & Context window saturation drift       & Avg.\ context utilisation $>$85\% \\
\midrule
\multirow{4}{*}{C2 Infra}
  & 2a & P99 latency regression                & Latency SLO breach \\
  & 2b & OOM / KV-cache exhaustion             & GPU memory $>$95\%; 5xx rate \\
  & 2c & Routing misclassification under load  & Wrong model serving traffic \\
  & 2d & API breaking change                   & Response schema mismatch \\
\midrule
\multirow{5}{*}{C3 Integration}
  & 3a & Prompt injection / jailbreak          & System prompt integrity check \\
  & 3b & Context truncation (silent)           & Context utilisation $>$90\% \\
  & 3c & Tool-call hallucination               & Tool-call success rate \\
  & 3d & RAG retrieval mismatch (stale index)  & Retrieval freshness timestamp \\
  & 3e & Multi-turn state corruption           & Per-session coherence score \\
\midrule
\multirow{4}{*}{C4 Evaluation}
  & 4a & Metric proxy collapse / benchmark gaming & Benchmark--production gap \\
  & 4b & Evaluation set contamination           & Training/eval overlap rate \\
  & 4c & Production--evaluation distribution gap & Covariate shift score \\
  & 4d & Point accuracy vs.\ distributional     & Per-cohort metric decomposition \\
\midrule
\multirow{5}{*}{C5 Safety}
  & 5a & PII leakage / memorisation             & PII regex on outputs \\
  & 5b & Hallucinated legal/medical citations   & Citation verification pipeline \\
  & 5c & Policy boundary violation              & Output policy classifier \\
  & 5d & Auditability gap                       & Audit log completeness \\
  & 5e & Copyright reproduction                 & Similarity to training corpus \\
\midrule
\multirow{4}{*}{C6 Operational}
  & 6a & Monitoring blind-spot                  & Alert coverage matrix \\
  & 6b & Runbook absence                        & Runbook inventory \\
  & 6c & Escalation breakdown                   & Time to escalation vs.\ SLO \\
  & 6d & Canary / shadow test gap               & Shadow traffic coverage \\
\bottomrule
\end{tabular}
\end{table}

\clearpage
\section{Dataset Sample}
\label{appendix:dataset}

Table~\ref{tab:dataset_sample} provides two representative incidents per
failure class.  Full 50-incident dataset with source URLs at
\url{https://github.com/priyanka25aug/llm-failure-taxonomy/blob/main/data/public_incidents/incidents.csv}.

\begin{table}[ht]
\caption{Representative incident sample (2 per class).}
\label{tab:dataset_sample}
\centering
\small
\resizebox{\linewidth}{!}{%
\begin{tabular}{llL{4.5cm}lll}
\toprule
\textbf{ID} & \textbf{Class} & \textbf{Incident} & \textbf{Sev.} &
\textbf{Detect.} & \textbf{MTTD} \\
\midrule
FTX-0001 & C1a & GPT-4 silent behaviour change (Mar 2023) & High & Silent & 168h \\
FTX-0002 & C1b & Production query distribution shift post-launch & Medium & Silent & 336h \\
\midrule
FTX-0007 & C2b & vLLM KV-cache exhaustion under bursty load & High & Immediate & 0.5h \\
FTX-0008 & C2a & ChatGPT outage latency spike (Dec 2022) & High & Immediate & 1h \\
\midrule
FTX-0013 & C3a & Bing Chat Sydney prompt injection (Feb 2023) & Critical & Delayed & 24h \\
FTX-0014 & C3d & RAG stale index serving outdated Basel III guidance & Critical & Silent & 2880h \\
\midrule
FTX-0026 & C4a & MMLU fine-tuned model degraded real-world performance & Medium & Silent & 720h \\
FTX-0027 & C4b & Evaluation set contamination in benchmark study & Medium & Silent & 1440h \\
\midrule
FTX-0031 & C5b & Mata v.\ Avianca hallucinated citations (2023) & Critical & Delayed & 48h \\
FTX-0032 & C5a & Samsung employee PII leak via ChatGPT (2023) & Critical & Delayed & 72h \\
\midrule
FTX-0044 & C6a & Output-length drift unmonitored for 3 weeks & Medium & Delayed & 504h \\
FTX-0045 & C6b & No runbook for model API deprecation scenario & High & Delayed & 120h \\
\bottomrule
\end{tabular}}
\end{table}

\clearpage
\section{Failure Budget Calculator}
\label{appendix:calculator}

The reference implementation computes budget utilisation per use case.
Core logic (Python pseudocode):

\begin{minipage}{\columnwidth}
\begin{Verbatim}[fontsize=\small]
BUDGET_RATES = {FC_A: 1.0, FC_B: 5.0,
                FC_C: 20.0, FC_D: 100.0}
SEVERITY_WEIGHTS = {critical: 3, high: 2,
                    medium: 1, low: 0.5}

def budget_utilisation(uc, failures):
  weighted = sum(SEVERITY_WEIGHTS[f.severity]
                 for f in failures)
  rate = weighted / (uc.requests / 1000)
  return (rate / BUDGET_RATES[uc.risk]) * 100
\end{Verbatim}
\end{minipage}

\vspace{0.5em}
Full implementation with reporting and multi-use-case portfolio tracking:
{\small\url{https://github.com/priyanka25aug/llm-failure-taxonomy/blob/main/src/failure_budget/calculator.py}}.

% =============================================================================
\section*{Acknowledgments}
% =============================================================================

The independent annotation study reported in Section~\ref{sec:dataset} was
conducted by Mukund Pandey (Staff ML Engineer, Meta), who is also the author
of \citet{pandey2026evaluating}, cited in this work.  The author declares this
relationship in the interest of transparency; the annotation task was completed
prior to any discussion of this paper's taxonomy definitions, and annotator
independence was maintained throughout.

\end{document}